\documentclass{article}
\usepackage{ijcai25}

\usepackage{times}
\usepackage{soul}
\usepackage{url}
\usepackage[hidelinks]{hyperref}
\usepackage[utf8]{inputenc}

\usepackage[small]{caption}
\usepackage{caption}
\DeclareCaptionFont{ninept}{\fontsize{9pt}{10pt}\selectfont}
\usepackage{graphicx}
\usepackage{amsmath}
\usepackage{amsthm}
\usepackage{booktabs}
\usepackage{algorithm}
\usepackage{algorithmic}
\usepackage[switch]{lineno}
\usepackage{color}

\usepackage{amssymb,amsfonts}

\def\mathbi#1{\textbf{\em #1}}

\newcommand{\eg}{\emph{e.g.}}

\usepackage{latexsym}

\usepackage{eso-pic}

\AddToShipoutPictureBG{%
  \AtPageUpperLeft{%
    \raisebox{-2.5\baselineskip}{%
      \hspace{1in} 
      \parbox{\textwidth}{%
        \centering
        \small Proceedings of the Thirty-Fourth International Joint Conference on Artificial Intelligence (IJCAI-25)\\
        Special Track on AI4Tech: AI Enabling Critical Technologies \\  ------------------------------ This is a Preprint (Accepted for publication at IJCAI2025 ------------------------------
      }
    }
  }
}

\title{DeepFeatIoT: Unifying Deep Learned, Randomized, and LLM Features for Enhanced IoT Time Series Sensor Data Classification in Smart Industries}

\author{
Muhammad Sakib Khan Inan
\And
Kewen Liao\footnote{Corresponding Author}
\\
\affiliations
School of Information Technology, Deakin University, Australia\\
\emails
sakib.inan@research.deakin.edu.au,
kewen.liao@deakin.edu.au
}

\begin{document}

\maketitle

\begin{abstract}
Internet of Things (IoT) sensors are ubiquitous technologies deployed across smart cities, industrial sites, and healthcare systems. They continuously generate time series data that enable advanced analytics and automation in industries. However, challenges such as the loss or ambiguity of sensor metadata, heterogeneity in data sources, varying sampling frequencies, inconsistent units of measurement, and irregular timestamps make raw IoT time series data difficult to interpret, undermining the effectiveness of smart systems. To address these challenges, we propose a novel deep learning model, DeepFeatIoT, which integrates learned local and global features with non-learned randomized convolutional kernel-based features and features from large language models (LLMs). This straightforward yet unique fusion of diverse learned and non-learned features significantly enhances IoT time series sensor data classification, even in scenarios with limited labeled data. Our model's effectiveness is demonstrated through its consistent and generalized performance across multiple real-world IoT sensor datasets from diverse critical application domains, outperforming state-of-the-art benchmark models. These results highlight DeepFeatIoT's potential to drive significant advancements in IoT analytics and support the development of next-generation smart systems.

\end{abstract}

\section{Introduction}
In the era of Industry 4.0, the world around us is surrounded by a vast number of IoT sensors capturing data from diverse real-world environments, enabling smart automated actions or decision-making across various critical sectors, including smart cities, smart buildings, smart agriculture, and smart healthcare. These sensors record a wide range of real-world observations such as temperature, humidity, traffic flow, snow height, visibility, and more \cite{mace_paper,Senshamart}.  A recent report from DemandSage\footnote{https://www.demandsage.com/number-of-iot-devices/} predicts that the number of connected IoT devices will reach 40 billion by 2030. The vast amount of time series sensor data generated from these devices fuels essential smart industrial applications, such as automated traffic monitoring in cities, environmental monitoring for sustainability, smart building management, airport operations, and precision agriculture. These advancements are driven by the advanced automatic analysis of historical time series sensor data \cite{wajgi2023internet,manoharan2021artificial,rajak2021internet}.

With the growing number of connected IoT devices, metadata loss, which refers to the supporting textual data that describes sensors and their observations, has become a common issue \cite{vculic2023lost}. This can result from network communication failures, battery depletion, a lack of standardized data storage and sharing systems \cite{vculic2023lost,mace_paper,inan2023deepheteroiot}, or, in some cases, metadata may not be stored at all due to security concerns \cite{stoyanova2020survey}. The absence of accurate metadata poses significant challenges for researchers and developers, rendering large volumes of historical IoT time series sensor data uninterpretable \cite{vculic2023lost,inan2023deepheteroiot}. This, in turn, undermines automated data analysis and smart decision-making processes, which are essential for a wide range of critical industrial applications. In such cases, we are left with only time series sensor readings from IoT devices. Before conducting any data analysis, we must first determine the specific type of IoT sensor (\eg, temperature, humidity, traffic flow) to which each time series data stream corresponds. This necessity gives rise to the challenge of IoT time series sensor data classification. Furthermore, classifying IoT time series sensor data is highly challenging due to its inherent complexity, which arises from sensors generating data from heterogeneous sources, various geographic locations, different sampling frequencies, and irregular timestamps \cite{mace_paper,vculic2023lost}. Additionally, manual, human-involved IoT data annotation or classification is time-consuming, labor-intensive, and often financially infeasible due to its high costs \cite{mace_paper}.

To address these issues, several artificial intelligence (AI) researchers have developed state-of-the-art machine learning \cite{mace_paper,vculic2023lost} and deep learning algorithms \cite{inan2023deepheteroiot} for the effective classification of IoT time series sensor data. Previously, an ensemble of multiple machine learning methods, using improved model selection and class filtering strategies, was found to outperform traditional time series classification methods for IoT sensor data \cite{mace_paper}. Although ensemble machine learning showed promising performance, it struggled to fully capture the heterogeneity in time series patterns (a mixture of local sub-patterns and global patterns) due to its reliance on pre-computed statistical properties from IoT time series data. More recently, deep learning algorithms \cite{inan2023deepheteroiot} have been introduced to learn complex local and global features directly from raw IoT time series data to facilitate classification. However, these models still face challenges in generalizing well when labeled data is limited in heterogeneous IoT sensor datasets. This limitation arises because deep learning algorithms typically require large amounts of training data, which is not always readily available in IoT sensor data classification.

Inspired by the recent success of large language models (LLMs) like GPT (Generative Pretrained Transformer) in Natural Language Processing \cite{min2023recent} and Computer Vision \cite{khan2022transformers}, as well as the effectiveness of randomized convolutions in capturing diverse time series patterns \cite{dempster2020rocket,tan2022multirocket}, we propose a novel deep learning model for IoT time series sensor classification. Our approach unifies learned multi-scale convolutional and recurrent representations with non-learned randomized convolutional features and pre-trained LLM features to enhance model generalization across heterogeneous IoT sensor datasets. The main contributions of this work are summarized as follows:

\begin{itemize}  
    \item  We propose a novel deep learning model, DeepFeatIoT, which uniquely integrates learned, randomized, and pre-trained LLM features to enhance classification performance across various IoT sensor datasets, even in scenarios with limited labeled data.
    
    \item We employ a uniform dense feature transformation module to fuse feature representations of different scales, ensuring balanced feature contributions. Through this module, our randomized convolutional features and pre-trained LLM features (\eg, from GPT-2~\cite{GPT2}) are effectively incorporated.
    
    \item We establish a comprehensive benchmark through rigorous empirical evaluations against state-of-the-art deep learning models across multiple real-world IoT sensor datasets, highlighting the importance of feature diversity for improved classification.
    
    
    
\end{itemize}  

To the best of our knowledge, we are the first to effectively fuse a range of learned and non-learned features within a deep learning model for classifying heterogeneous IoT sensor types, supporting critical applications in smart industries.

\label{sec:intro}
\section{Related Works}
This section is organized into two subsections: \textit{IoT Time Series Sensor Data Classification} and \textit{Fundamental Time Series Classification}. The first subsection reviews state-of-the-art methods specifically designed for IoT time series sensor data, while the second highlights recent algorithmic advancements in general time series classification. 

\label{sec:related-works}
\subsection{IoT Time Series Sensor Data Classification}

IoT time series sensor data are often noisy, with highly correlated patterns across class labels, overlapping sub-patterns, and inherent heterogeneity due to variations in timestamp ranges, sampling ratios, frequencies, or units of measurement. This makes the classification of IoT sensors from time series data a challenging task. Additionally, irregularities in time series intervals, unlike traditional time series classification problems where data is equally spaced in time, further complicate the process. Due to the popularity of statistical transformation and data mining methods in the time series classification domain, researchers previously developed a probabilistic data mining approach \cite{Swiss-Experiment} that incorporated slope distribution computation via linear approximation of IoT time series sequences for heterogeneous IoT sensor data classification. Postol et al.~\cite{postol2019time} later proposed a topological data analysis-based strategy for noisy IoT sensor data classification, while Borges et al.~\cite{borges2022classification} introduced a transformation-based classification strategy that converts raw sensor data into ordinal patterns, improving feature representation and class separability. However, statistical transformation methods often struggle due to their assumptions about data distribution, especially with the variability and non-linearity of time series patterns, which may involve a mix of local and global patterns. Although these methods may work well for specific datasets, recent studies have found them to be ineffective across a variety of heterogeneous IoT sensor data classification scenarios \cite{mace_paper}. Recently, ensemble machine learning has outperformed statistical transformation and data mining methods, offering better generalizability, as demonstrated by Montori et al.~\cite{mace_paper} and Gambiroza et al.~\cite{vculic2023lost}. However, despite their success, ensemble machine learning methods often suffer from overfitting in many use cases. Very recently, deep learning has emerged as a promising approach, leveraging the ability to learn global and local patterns in parallel to improve classification performance \cite{inan2023deepheteroiot}. Deep learning algorithms, however, require specific pre-processing or augmentations in cases where the number of samples is limited, particularly when there is class imbalance. This could potentially lead to adding bias or cause data-leakage.

\subsection{Fundamental Time Series Classification}
In fundamental time series classification problems, data points are typically equally spaced over time, and the dataset generally contains time series from the same or similar domains or applications, which is often different in the case of IoT time series \cite{bagnall2017great,vculic2023lost}. Over the years, fundamental time series classification methods have seen significant growth and advancement. It began with classical distance-based methods like Dynamic Time Warping (DTW) \cite{bagnall2017great} and has evolved to more recent methods, such as the Large Language Model-based approach GPT4TS~\cite{OFA-LLM}. Deep learning in time series classification gained attention when Ismail et al.\cite{ismail2020inceptiontime} introduced the state-of-the-art convolutional architecture InceptionTime, inspired by AlexNet \cite{alexnet}. Recently, randomized convolutions became popular in the time series classification space with the introduction of scalable random convolutional kernel-based methods like ROCKET \cite{dempster2020rocket} and its extension MultiROCKET \cite{tan2022multirocket}. Self-attention-based transformer architectures have also shown promising performance, particularly with the introduction of the patching concept for time series, proving the effectiveness of dividing a time series sequence into 64 patches for use in the transformer architecture named PatchTST \cite{PatchTST}. Later, the researchers of PatchTSMixer~\cite{PatchTSMixer} developed a Multi-Layer Perceptron (MLP)-based approach using the patching concept over time series sequences, making it more lightweight and faster than previous self-attention-dominant methods like PatchTST~\cite{PatchTST}. The growing popularity of large language models (LLM) has also inspired AI researchers in the time series domain to develop a Generative Pre-trained Transformer (GPT)-based method, GPT4TS \cite{OFA-LLM}, which has achieved competitive performance in fundamental time series tasks.

While these methods perform well in fundamental time series classification problems, due to the unique nature of IoT time series data, similar methods have been found to struggle or be ineffective in a wide range of heterogeneous IoT time series classification problems \cite{inan2023deepheteroiot,mace_paper}. Although there have been recent developments in deep learning algorithms specifically designed for IoT time series data classification, as previously discussed, they still suffer from overfitting when faced with a limited number of labelled samples along with class imbalance~\cite{inan2023deepheteroiot}. Thus, a research gap remains in advancing state-of-the-art methods for improved feature extraction to support the classification of IoT time series sensor data, despite the limited labelled data challenges present in the data.
\section{Method: DeepFeatIoT}
\begin{figure*}[tb]
    \centering
    \includegraphics[width=0.9\linewidth]{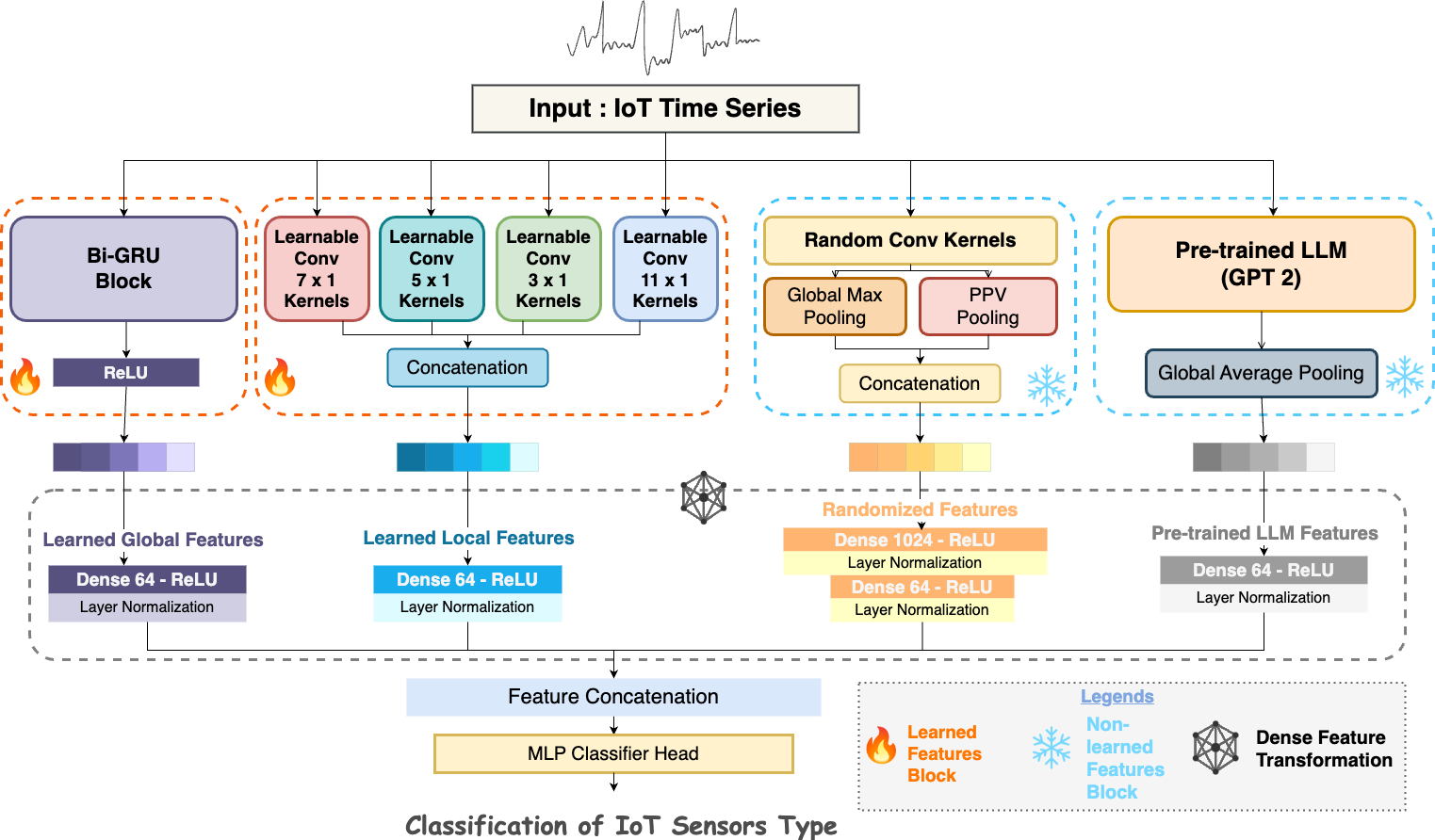}
    \caption{Illustration of the proposed deep learning model DeepFeatIoT.}
    \label{fig:proposed-model}
\end{figure*}

In this section, we delineate the details of our proposed deep learning model, DeepFeatIoT. An overview of DeepFeatIoT is illustrated in Figure \ref{fig:proposed-model}. DeepFeatIoT takes raw IoT time series sensor data as input (without any preprocessing) and extracts deep-learned local and global features, randomized convolutional features, and pre-trained LLM features. These four diverse sets of latent feature representations are then combined through an optimized dense feature transformation (DFT) module, leading to improved classification of IoT time series sensor data. The following subsections provide a detailed methodology of the proposed DeepFeatIoT model.
\label{sec:method}

\subsection{Preliminaries}

An individual sample of IoT sensor data from the whole dataset is essentially a vector containing a sequence of real numbers, denoted as $\mathbi{X}_{i} = (x_{1}, x_{2}, x_{3}, \dots, x_{n})$, with $i$ indexing the $i$-th sample from the dataset and $n$ representing the number of timestamps. Each $x_{j}$ denotes a numerical value of sensor reading at the $j$\textit{th} timestamp. Then the task of IoT sensor data classification can be formulated as determining the value of $y_{i}$, which is the designated class label (\eg, temperature, humidity, light sensor, etc.) for the $i$\textit{th} sample $\mathbi{X}_{i}$ (time series sequence) from the dataset.

\subsection{Learned Local and Global Features}
As IoT sensor data generates from heterogeneous sources and the datasets contain sensors from diverse domains, it is important to extract both global pattern and local sub-pattern from IoT time series for better feature representation. In this regard, we adopted a learned feature extraction strategy proposed in a previous study \cite{inan2023deepheteroiot} with some minor modification. In our proposed model, we incorporated a stack of bi-directional gated recurrent units (Bi-GRU) layers to extract deep learned global features and a stack of learned non-dilated convolutional (conv) kernels (conv layers with different kernel sizes) to extract deep learned local features. Moreover, in the learned global feature extraction process, to mitigate the vanishing gradient problem in recurrent neural networks \cite{vanishing-gradient}, we applied the ReLU (Rectified Linear Unit) transformation to clip negative values in the global learned latent feature representation. This learned feature extraction process can be mathematically represented as, 
\begin{equation}\label{eg:global-feature}
    \mathcal{{F}}_{g} = \sigma(G(\mathbi{X}_{i})),
\end{equation}
where $\mathbi{X}_{i}$ is the raw IoT time series sequence as describe earlier, $G$ is the function for computing global feature using Bi-GRU layers and $\sigma$ is the ReLU activation, then the resultant is global learned feature vector $\mathcal{{F}}_{g}$.
On the other hand, features from multi-scale convolutional blocks is concatenated along the depth dimension, which can be similarly represented mathematically as:
\begin{equation}\label{eq:local-feature}
    \mathcal{{F}}_{c} =\mathcal{K}_{3}(\mathbi{X}_{i}) || \mathcal{K}_{5}(\mathbi{X}_{i}) || \mathcal{K}_{7}(\mathbi{X}_{i}) || \mathcal{K}_{11}(\mathbi{X}_{i}),
\end{equation}
where $\mathcal{K}_{i}$ represents function with stack of convolutional layers of kernel size $i$ and $\mathcal{{F}}_{c}$ represents latent feature vector comprising of deep learned local features.
\subsection{Randomized Features}
Randomized non-learned convolutional kernels based feature transform of input time series were proven to effective for capturing discriminating features for fundamental time series classification by several previous studies \cite{dempster2020rocket,jimenez2019time}. Large amount of randomized convolutional kernel transformation can produce variety of diverse non-linear features that could be effectively used for further classification task. This method is scalable and works well for small time series datasets with limited sample and labels, effecitvely avoiding overffiting due to its unsupervised nature in feature extraction stage \cite{dempster2020rocket}. In this regard, considering the randomness in temporal pattern for IoT time series sensor data due to inherent heterogeneity, noise and data sparsity issue, randomized convolutional kernels could be a scalable solution to capture diverse features with different aspects. So, we incorporate ten thousand randomized convolutional kernel over raw IoT time series sequence ($\mathbi{X}_{i}$) inspired by the works of \cite{dempster2020rocket} with modified and improved parameter configurations. In our proposed model, for extraction of randomized features, the kernel weights ($\mathbi{w}_{i}$) of the randomized kernels ($C_{i}$) were drawn from the normal distribution of $\mathcal{N}(0, 0.05)$. The length ($\mathbi{k}_{i}$)  of all kernels was set to 9. Unlike from previous work \cite{dempster2020rocket}, we set the bias ($ \mathbi{b}_{i}$) term  to zero and use a fixed dilation ($\mathbi{d}_{i}$) size 4. These deterministic modifications of parameter values were made to make the model more generalized across IoT time series sensor data. This process can be represented as:
\begin{equation}\label{eq:random-feature}
    \mathcal{{F}}_{r} = \sum_{i=1}^{m} C_{i}\{\mathbi{w}_{i}, \mathbi{k}_{i}, \mathbi{b}_{i}, \mathbi{d}_{i}\}(\mathbi{X}_{i}),
\end{equation}
where $\mathcal{{F}}_{r}$ represents latent feature vector of randomized convolution kernels (the number of kernels $m=10000$) aggregated via global max pooling and proportion of positive value pooling. 
The updated parameter configuration is also more scalable due to its fixed dilation and padding, while the statistical distribution of weights has been reduced to a standard deviation of 0.05.

\subsection{Pre-trained LLM Features}
Pre-trained LLMs designed for solving natural language processing tasks, have already proven effective at capturing long-term sequential patterns from complex data and can be used to solve a variety of challenging problems~\cite{yang2024harnessing}. Especially, when there is limited availability of data and labels in downstream tasks. Time series data has a similar structure to textual data \cite{gruver2024large}, as both text and time series data exhibit sequential nature and correlation. The main difference is that, in time series data, the elements of a sequence are real-valued numbers, whereas in the text domain it is generally sequence of words in natural language \cite{gruver2024large}.  Inspired by the works of \cite{OFA-LLM}, to utilize the benefit of pre-trained LLM features in the context of IoT time series sensor data, we incorporate pre-trained GPT2~\cite{GPT2} in our proposed model for extracting LLM based sequential contextual features from raw IoT time series sequence. Unlike the previous work \cite{OFA-LLM}, in the proposed model, we do not do any re-programming or transformation of raw IoT time series, instead, a IoT time series sequence $\mathbi{X}_{i}$, is tokenized as textual sentence (numbers treated as sequence of characters or words) where each element or value, $x_{j}$ at individual timestamp is separated by special characters to make up a text sequence and fed as input to the GPT2~\cite{GPT2} model after tokenization. We incorporate all the 12 layers (transformer blocks) of GPT2 in it's original form. This would help the model to take benefit of pre-training on text data and extract sequential correlation in IoT time series without any transformation or re-programming. This can be mathematically represented as
\begin{equation}\label{eq:transefrrable-llm-feature}
    \mathcal{F}_{l} = pool_{avg}(GPT2(\{\mathbi{X}_{i}\}))\,.
\end{equation}
Here, a two-dimensional feature vector is generated as output by the GPT2 model which goes through a Global Average Pooling layer ($pool_{avg}$) to produce one-dimensional latent features space vector $\mathcal{F}_{l}$, that represents our proposed pre-trained LLM features capturing sequential contextual correlation in raw IoT time series data.
\subsection{Dense Feature Transformation}
After extracting the learned latent feature vectors $\mathcal{F}_{g}$ and $\mathcal{F}_{c}$, along with the randomized features $\mathcal{F}_{r}$ and pre-trained LLM features $\mathcal{F}_{l}$, the next step is to optimally combine these four diverse features into a single latent space to facilitate the classification step. These feature vectors vary in dimensionality or feature space size. For instance, the dimensionality of $\mathcal{F}_{r}$ and $\mathcal{F}_{l}$ are $\mathcal{F}_{r} \in \mathbb{R}^{20000 \times 1}$ and $\mathcal{F}_{l} \in \mathbb{R}^{768 \times 1}$, which are much larger compared to the dimensionalities of $\mathcal{F}_{g}$ and $\mathcal{F}_{c}$, which are $\mathcal{F}_{g} \in \mathbb{R}^{128 \times 1}$ and $\mathcal{F}_{c} \in \mathbb{R}^{256 \times 1}$, respectively.

At this stage, directly concatenating these feature vectors with varying dimensions as input to the next fully connected feed-forward layer (FCN) would result in bias where the larger and sparser feature vectors would dominate over smaller feature vectors in terms of dimensions. This could lead to overfitting, poor generalization, and suffering from the curse of dimensionality. To mitigate these issues, we designed the Dense Feature Transformation (DFT) module, a simple yet effective method for transforming each feature vector into a dense vector space with reduced dimensionality. This acts as an indirect step for the selection and scaling of features within the neural network. The DFT module transforms the four different feature vectors into separate dense vector spaces of equal size (64 dimensions), addressing the issues mentioned above. It can be represented in simplified mathematical form as:
\begin{equation}
    \label{eq:DFT}
   \resizebox{.91\linewidth}{!}{$
   \displaystyle
   \mathcal{F}_{d} = \sigma(W_{g}\{\mathcal{F}_{g}\}) || \sigma(W_{c}\{\mathcal{F}_{c}\}) \\ || \sigma(W_{r}\{\mathcal{F}_{r}\}) || \sigma(W_{l}\{\mathcal{F}_{l}\}) , $}
\end{equation}
where $\sigma$ represents the non-linear activation function, which in this case is ReLU, and $W_{g}, W_{c}, W_{r}, W_{l}$ are the weight matrices applied respectively to $\mathcal{F}_{g}, \mathcal{F}_{c}, \mathcal{F}_{r}, \mathcal{F}_{l}$. Among these, $W_{g}, W_{c},$ and $W_{l}$ are dense layers with a size of $\mathbb{R}^{64}$, while $W_{r}$ consists of a stack of two dense layers with sizes $\mathbb{R}^{1024}$ and $\mathbb{R}^{64}$, respectively. Each dense fully connected layer is followed by a layer normalization operation. The output of DFT module is $\mathcal{F}_{d}$ a combined feature vector containing all 4 types features represented in a single latent space.
\begin{algorithm}[t]
\caption{DeepFeatIoT Algorithmic Pseudo-code}
	\label{algo:deepfeatiot}
    	\hspace*{\algorithmicindent}{\textbf{Input:} $\mathbi{X}_{i}$ represents a particular IoT time series sequence from the IoT sensor dataset, which can be mapped to a particular class label $y_{i}$. So, a IoT sensor dataset can be represented as $\mathcal{D} = \{(\mathbi{X}_{i},y_{i}), \dots, \dots,(\mathbi{X}_{n},y_{n})\}$. For a classification problem, the model is going be trained iterating over every $i$th instance in the dataset,  $\mathcal{D}$ ($n$ represents number of samples/instances in the dataset).
	
 \begin{algorithmic}[1]
	\FOR{Each $epoch$ in $Epochs$}
        \STATE For each $i$ th instance ($\mathbi{X}_{i}$) compute learned features $\mathcal{F}_{g}$ and $\mathcal{F}_{c}$ using accordingly equations \eqref{eg:global-feature} and \eqref{eq:local-feature}.
        \STATE Similarly, compute randomized convolutional kernel feature $\mathcal{F}_{c}$ using equation \eqref{eq:random-feature} and extract pre-trained LLM features, $\mathcal{F}_{l}$, from LLM using equation \eqref{eq:transefrrable-llm-feature}.
        \STATE Combine features $\mathcal{F}_{g}$, $\mathcal{F}_{c}$, $\mathcal{F}_{r}$, $\mathcal{F}_{l}$ into a single feature vector space $\mathcal{F}_{d}$ using \eqref{eq:DFT}.
	  \STATE $\mathcal{F}_{d}$ goes into MLP Head and through softmax layer and generate probability distribution to determine the value of $y_{i}$.
        \STATE Compute errors in prediction after each batch and update weights.
	\ENDFOR
	\end{algorithmic}
	\hspace*{\algorithmicindent}{\textbf{Output:} A trained deep learning model}}
\end{algorithm}
After that, as illustrated in Figure \ref{fig:proposed-model}, the combined feature vector is fed into an MLP head, which consists of a stack of two feed-forward neural network layers with 128 and 64 neurons, respectively. Each layer is followed by layer normalization and a dropout of 50\%. Finally, the output from the last hidden layer is passed through a softmax layer for the classification task.. Algorithm \ref{algo:deepfeatiot} presents the pseudo-code of the proposed DeepFeatIoT algorithm, aligning with the brief description of the individual modules outlined above. The proposed DeepFeatIoT model incorporates a wide range of learned and non-learned features to capture diverse contextual aspects from raw IoT time series, enhancing generalizability across various IoT sensor data domains, even in scenarios with limited data and label availability, without requiring any data pre-processing. The DeepFeatIoT model is trained for 200 epochs, and the best weights, based on testing accuracy scores, are used for the final experimental evaluation. The training is conducted using the popular Adam optimizer with an initial learning rate of 0.001, combined with inverse time decay every 100 steps, and categorical focal cross-entropy loss.

\begin{table}[b]
\centering

\begin{tabular}{@{}ccccc@{}}
\toprule
\textbf{Dataset} & \textbf{Length} & \textbf{Samples} & \textbf{Classes} & \textbf{Domain} \\ \midrule
Swiss & 445 & 346 & 11 & Smart City \\
Urban & 864 & 1065 & 16 & Smart City \\
Iowa & 168 & 1000 & 8 & Smart Airport \\
SBAS & 168 & 255 & 5 & Smart Buildings \\ \bottomrule
\end{tabular}
\caption{A summary of IoT time series datasets and respective critical industrial application domains}
\label{tab:dataset-summary}
\end{table}

\begin{table*}[tb]
\centering

\begin{tabular}{@{}rcccccccccc@{}}
\toprule
 & \multicolumn{2}{c}{Swiss} & \multicolumn{2}{c}{Urban} & \multicolumn{2}{c}{Iowa} & \multicolumn{2}{c}{SBAS} & \multicolumn{2}{c}{Average} \\ \midrule
 & Acc & F1 & Acc & F1 & Acc & F1 & Acc & F1 & Acc & F1 \\
InceptionTime & 67.31 & 58.93 & 90.31 & 80.76 & \color{blue}\underline{96.67} & \color{blue}\underline{96.70} & 97.40 & 97.42 & 87.92 & 83.45 \\
H-InceptionTime & 75.00 & 62.66 & 90.94 & 81.01 & \color{blue}\underline{96.67} &\color{blue}\underline{96.70} & 97.40 & 97.41 & 90.00 & 84.45 \\
PatchTST & 62.50 & 56.86 & 85.63 & 74.54 & 87.99 & 87.96 & 89.62 & 89.50 & 81.44 & 77.22 \\
PatchTSMixer & 71.15 & 57.41 & 90.31 & 84.80 & 92.00 & 92.02 & \color{red}\textbf{100.00} & \color{red}\textbf{100.00} & 88.37 & 83.56 \\
GPT4TS & 70.19 & 62.54 & 86.56 & 80.98 & 88.33 & 88.22 & \color{red}\textbf{100.00} & \color{red}\textbf{100.00} & 86.27 & 82.94 \\
ROCKET & 75.00 & 72.13 & 90.31 & 82.81 & 91.00 & 91.02 & 97.40 & 97.48 & 88.43 & 85.86 \\
MultiROCKET & 84.62 & 82.76 & 96.56 & \color{red}\textbf{95.23} & 95.00 & 94.96 & \color{blue}\underline{98.70} & \color{blue}\underline{98.74} & 93.72 & 92.92 \\
MACE & 86.54 & \color{blue}\underline{84.01} & 92.25 & 88.74 & 91.01 & 91.04 & 97.40 & 97.49 & 91.80 & 90.32 \\
DeepHeteroIoT & \color{blue}\underline{86.54} & 82.92 & \color{blue}\underline{96.88} & 94.44 & 96.01 & 96.05 & \color{red}\textbf{100.00} & \color{red}\textbf{100.00} & \color{blue}\underline{94.86} & \color{blue}\underline{93.35} \\
\textbf{DeepFeatIoT} & \color{red}\textbf{91.35} & \color{red}\textbf{90.98} & \color{red}\textbf{97.19} & \color{blue}\underline{94.79} & \color{red}\textbf{99.33} & \color{red}\textbf{99.33} & \color{red}\textbf{100.00} & \color{red}\textbf{100.00} & \color{red}\textbf{96.97} & \color{red}\textbf{96.28}\\ \bottomrule
\end{tabular}
\caption{Comparison of classification accuracy (acc) and macro-average F1 (F1) score  of proposed model against state-of-the art models across IoT Sensor datasets. \color{red}{\textbf{Bold}}: best results, \color{blue}{\underline{Underline}}: second best results}
\label{tab:performance-comp}
\end{table*}

\section{Experimental Results}
\label{sec:experimental-results}

This section discusses the experimental designs and validations of our proposed model across IoT sensor datasets containing time series sensor data serving various critical applications and domains. It also presents a benchmark analysis against previous state-of-the-art studies. Each IoT sensor dataset was resampled (stratified) into a training and testing ratio of 70:30 (random seed 100) \cite{mace_paper}. The detailed programming or source code implementation of this research can be found at the following GitHub repository: \href{https://github.com/skinan/DeepFeatIoT-IJCAI-2025}{{\textit{https://github.com/skinan/DeepFeatIoT-IJCAI-2025}}}

Furthermore, to establish robust benchmark performance comparison baselines, we include state-of-the-art IoT sensor classification models such as MACE \cite{mace_paper} and DeepHeteroIoT \cite{inan2023deepheteroiot}, as well as advanced time series classification models, including convolution-based InceptionTime~\cite{ismail2020inceptiontime}, and H-InceptionTime~\cite{H-InceptionTime}, transformer-based PatchTST~\cite{PatchTST}, MLP-mixer-based PatchTSMixer~\cite{PatchTSMixer}, LLM-based GPT4TS~\cite{OFA-LLM}, randomized convolution based ROCKET~\cite{dempster2020rocket} and MultiROCKET~\cite{tan2022multirocket} in the evaluation stage. As evaluation metrics, we consider two popular and standard evaluation metrics Accuracy and macro-average F1 score. For the fair comparison of performance evaluation, all the deep learning models are trained for 200 epochs with a learning rate of 0.001.

The IoT sensor datasets utilized in this study are: Swiss Experiment (Swiss) \cite{mace_paper}, Urban Observatory\footnote{http://newcastle.urbanobservatory.ac.uk/} (Urban) \cite{mace_paper},  Iowa ASOS\footnote{https://mesonet.agron.iastate.edu/ASOS/} (Iowa) \cite{inan2023deepheteroiot}, Smart Building Automation System (SBAS) \cite{hong2017high}. Among all these datasets, the Swiss dataset contains a high level of noise and heterogeneity along with the issue of class imbalance and a very limited number of labelled samples \cite{Swiss-Experiment,mace_paper}. The Urban dataset contains highly correlated IoT sensor data. 

On the other hand, Iowa and SBAS datasets are balanced in terms of number of samples per class. Table \ref{tab:dataset-summary} summarizes key information about sensor data sets, including the length of time series, the number of samples, the number of classes, and the domain of critical industrial applications.

\subsection{Classification Performance}

Table~\ref{tab:performance-comp} summarizes the classification Accuracy (Acc) and the corresponding macro-average F1 (F1) scores of our proposed model, DeepFeatIoT, compared with several state-of-the-art baselines evaluated on diverse IoT sensor datasets. The table reports the best results obtained from 10 independent runs for each model. Notably, DeepFeatIoT achieves a maximum accuracy of 96.97\% and an average F1 score of 96.28\%, thereby surpassing the second-best model, DeepHeteroIoT~\cite{inan2023deepheteroiot}, by more than 2\%. Overall, when considering both metrics, the top five ranked models are: DeepFeatIoT (proposed), DeepHeteroIoT~\cite{inan2023deepheteroiot}, MultiROCKET~\cite{tan2022multirocket}, MACE~\cite{mace_paper}, and H-InceptionTime~\cite{H-InceptionTime}, with average accuracies close to 90\% or higher.

DeepFeatIoT consistently outperforms all benchmark models in accuracy on an individual dataset basis. For instance, on the heterogeneous Swiss dataset, DeepFeatIoT attains an F1 score of 90.98, which is approximately 6\% higher than that of MACE~\cite{mace_paper} and about 8\% higher than those of both MultiROCKET~\cite{tan2022multirocket} and DeepHeteroIoT~\cite{inan2023deepheteroiot}. In the Urban dataset, although MultiROCKET~\cite{tan2022multirocket} achieves a marginally higher F1 score, DeepFeatIoT slightly outperforms it in accuracy. Notably, DeepHeteroIoT~\cite{inan2023deepheteroiot} also exhibits competitive performance on the Urban dataset, ranking second in accuracy and third in F1 score, which indicates a close competition among these models.  

In the Iowa dataset, DeepFeatIoT achieves an exceptional accuracy and F1 score of 99.33, outperforming the second-best models (InceptionTime~\cite{ismail2020inceptiontime} and H-InceptionTime~\cite{H-InceptionTime}) by nearly 2.6\%. It is worth emphasizing that while ensembles of convolution-based architectures (\eg, InceptionTime~\cite{ismail2020inceptiontime} and H-InceptionTime~\cite{H-InceptionTime}) can yield impressive results on well-balanced datasets, their performance often deteriorates on datasets with imbalanced class distributions (\eg, Swiss).

\begin{table}[b]
\centering

\begin{tabular}{@{}rcccc@{}}
\toprule
 & Swiss & Urban & Iowa & SBAS \\ \midrule
RF & 78.1$\pm$0.4 & 87.3$\pm$0.4 & 83.3$\pm$1.2 & 97.4$\pm$0.0 \\
PF & 78.6$\pm$1.1 & 90.8$\pm$0.5 & 88.6$\pm$0.4 & 99.9$\pm$0.1 \\
RF \& PF & 78.8$\pm$0.8 & 86.2$\pm$0.5 & 88.2$\pm$1.5 & 98.7$\pm$0.0 \\
DC & 77.4$\pm$1.2 & 90.3$\pm$1.3 & 83.2$\pm$3.3 & 98.7$\pm$0.0 \\
Ours & 89.7$\pm$0.9 & 96.5$\pm$0.3 & 98.7$\pm$0.4 & 98.9$\pm$0.5 \\ \bottomrule
\end{tabular}
\caption{Ablation analysis of our proposed model based on mean and standard deviation of accuracy scores}
\label{tab:ablation-accuracy}
\end{table}

In summary, our experimental results indicate that DeepFeatIoT is the most effective model when evaluated across four diverse IoT sensor datasets, as evidenced by its superior average accuracy and F1 scores. To assess efficiency, the average runtime (training and testing) of DeepFeatIoT  across all datasets is approximately 10.8 minutes, which is commendable given its intricate architecture combining learned and non-learned feature extractors, yet remaining feasible for real-world use. Benchmark models such as DeepHeteroIoT~\cite{inan2023deepheteroiot} and MultiROCKET~\cite{tan2022multirocket} are the next best alternatives in terms of consistency, while the previous state-of-the-art ensemble ML benchmark, MACE~\cite{mace_paper}, demonstrates competitive performance with an average accuracy of 91.80\% and a F1 score of 90.32\%.  In contrast, transformer-based models like PatchTST~\cite{PatchTST} perform poorly in the IoT time series classification domain, with average scores of 81.44\% accuracy and 77.22\% F1. Nevertheless, larger transformer architectures (\eg, GPT4TS~\cite{OFA-LLM}) that benefit from pre-training, exhibit promising potential by achieving perfect scores on the SBAS dataset and competitive overall averages. Moreover, DeepFeatIoT, PatchTSMixer~\cite{PatchTSMixer}, and DeepHeteroIoT~\cite{inan2023deepheteroiot} all achieve perfect scores (100\%) in both accuracy and F1 on the SBAS dataset.

\subsection{Ablation Study}
In this section, we present a detailed ablation study for major components of our proposed DeepFeatIoT model, to validate the effectiveness of our proposed deep learning model across datasets. Tables \ref{tab:ablation-accuracy} and \ref{tab:ablation-F1} respectively delineate the mean $\pm$ standard deviation of accuracy and F1 scores over 10 runs. In both tables, the ablation of the proposed model is broken down into four parts: using only randomized features (RF), using only pre-trained LLM features (PF), using both randomized features and pre-trained LLM features (RF \& PF), and finally, the results of direct concatenation (DC), which includes all learned and non-learned features but excludes our designed DFT module for improved feature transformation to scale the concatenation of dense features before the MLP classifier head.

\begin{table}[tb]
\centering
\begin{tabular}{@{}rcccc@{}}
\toprule
 & Swiss & Urban & Iowa & SBAS \\ \midrule
RF & 71.9$\pm$0.4 & 81.2$\pm$0.9 & 82.7$\pm$1.1 & 97.4$\pm$0.0 \\
PF & 72.6$\pm$2.4 & 81.1$\pm$2.0 & 87.8$\pm$0.9 & 99.9$\pm$0.1 \\
RF \& PF & 73.4$\pm$1.5 & 75.5$\pm$2.9 & 88.2$\pm$1.5 & 98.7$\pm$0.0 \\
DC & 71.1$\pm$1.8 & 82.3$\pm$3.2 & 82.8$\pm$3.9 & 98.7$\pm$0.0 \\
Ours & 88.7$\pm$1.2 & 93.9$\pm$0.4 & 98.7$\pm$0.4 & 98.9$\pm$0.5 \\ \bottomrule
\end{tabular}
\caption{Ablation analysis of our proposed model based on mean and standard deviation of macro-average F1 scores}
\label{tab:ablation-F1}
\end{table}

It is evident from the mean accuracy and F1 scores presented in Tables~\ref{tab:ablation-accuracy} and \ref{tab:ablation-F1} that our final proposed model which is an optimized unification of learned and non-learned features outperforms all other configurations.However, excluding the DFT module causes a significant drop in generalizability, as indicated by lower accuracy and F1 scores on three out of the four datasets. While configurations using only RF or only PF yield competitive performance, they do not consistently achieve superior results across all datasets. Furthermore, the combination of RF \& PF, when employed without the inclusion of learned local and global features and the DFT module, fails to deliver competitive performance.

To statistically quantify the practical significance of our model’s improvements, we computed Cohen’s \textit{d} statistics\footnote{Following Cohen’s guidelines \cite{cohen2013statistical}, effect sizes are interpreted as: Cohen's \textit{d} statistic $\geq 0.8$ (large, substantial improvement), indicating strong practical relevance of the observed performance gains.} \cite{cohen2013statistical} for mean accuracy and F1 score comparisons. The results indicate strong effect sizes across all ablation variants, with values exceeding 1.0 in all cases, reinforcing the substantial performance gains achieved by our proposed model. Specifically, DeepFeatIoT demonstrates a significantly higher generalization capability, with Cohen’s \textit{d} values reaching 1.44 (acc) and 1.43 (F1) against the RF-only variant, and consistently exceeding 1.0 across all comparisons. Overall, the computed Cohen’s \textit{d} effect sizes for mean accuracy indicate substantial performance differences: Ours vs RF (1.443), Ours vs PF (0.939), Ours vs RF \& PF (1.216), and Ours vs DC (1.191). Likewise, the effect sizes for F1 scores further reinforce this trend, with values of Ours vs RF (1.433), Ours vs PF (1.098), Ours vs RF \& PF (1.231), and Ours vs DC (1.299), highlighting the significant impact of our model’s unified feature representation.

Furthermore, these results validate that the unified representation of features of our model - particularly the unification of learned local and global features with randomized and pre-trained LLM features - plays a critical role in improving classification performance in diverse IoT sensor datasets. Furthermore, the large effect sizes (Cohen's \textit{d} $\geq 0.8$) confirm that the observed performance gains are not only statistically significant but also practically meaningful for real-world IoT sensor classification. Additionally, our experimental observations and analysis also suggest that state-of-the-art pre-trained LLM architectures originally designed for text generation, such as GPT2~\cite{GPT2}, can be effectively incorporated for feature extraction from raw IoT time series even without domain-specific adapter blocks like previous studies \cite{OFA-LLM} or re-programming \cite{PatchTST} or transformation (\eg, frequency domain conversion) of raw time series, paving the way for future research in heterogeneous IoT sensor data classification for critical application domains.

\section{Conclusion}
\label{sec:conclusion}
While previous state-of-the-art models often struggle with generalization or overfitting, particularly due to challenges like limited labelled data and class imbalance, our proposed deep learning model effectively addresses these issues without requiring additional data preprocessing or augmentation, even for smaller IoT sensor datasets (\eg, Swiss). The automated classification of IoT sensor data eliminates the need for manual annotation from heterogeneous sources, paving the way for smarter critical applications across various industries. If deployed in real-world industrial settings, the proposed model would significantly reduce manual labor while saving both financial costs and valuable time across multiple critical sectors. Ultimately, this approach promotes the reusability of vast amounts of meaningful IoT sensor data, enabling advancements in critical technologies.





\section*{Acknowledgments}
This research was funded by the Australian Research Council (ARC) through an ARC Discovery Project grant (DP220101420).


\bibliographystyle{named}

\end{document}